\documentclass[]{article}
\usepackage{proceed2e}

\usepackage{times}
\usepackage{amssymb}
\usepackage{amsmath}
\usepackage{float}
\usepackage{subfigure}
\usepackage{subcaption}
\usepackage{graphicx}
\usepackage{algorithm}
\usepackage{algorithmic}
\usepackage{multirow}
\usepackage{booktabs}
\usepackage[backend=biber, sorting=none, style=authoryear]{biblatex}
\usepackage{makecell}
\usepackage{stfloats}
\usepackage{tabularray}
\usepackage{enumitem}
\title{Outlier-Aware Training for Low-Bit Quantization of Structural Re-Parameterized Networks}

\author{Muqun Niu*, Yuan Ren*, Boyu Li and Chenchen Ding} 

\begin{document}

\maketitle

\begin{abstract}
Lightweight design of Convolutional Neural Networks (CNNs) requires co-design efforts in the model architectures and compression techniques. As a novel design paradigm that separates training and inference, a structural re-parameterized (SR) network such as the representative RepVGG revitalizes the simple VGG-like network with a high accuracy comparable to advanced and often more complicated networks. However, the merging process in SR networks introduces outliers into weights, making their distribution distinct from conventional networks and thus heightening difficulties in quantization. To address this, we propose an operator-level improvement for training called Outlier Aware Batch Normalization (OABN). Additionally, to meet the demands of limited bitwidths while upkeeping the inference accuracy, we develop a clustering-based non-uniform quantization framework for Quantization-Aware Training (QAT) named ClusterQAT. Integrating OABN with ClusterQAT, the quantized performance of RepVGG is largely enhanced, particularly when the bitwidth falls below 8.
\end{abstract}

\section{INTRODUCTION}
Convolutional Neural Networks (CNNs) play a major role in a multitude of computer vision tasks. To deploy CNNs onto edge devices with limited resources, lightweight design techniques are required to strike a balance between hardware cost and inference accuracy.

A great deal of literature demonstrates considerable efforts in lightweight design. Their contributions can be categorized into two main aspects – model architectural design and their compression algorithms. For the former, [\cite{he2016deep}] and [\cite{huang2017densely}] propose changes to the connection between Convolutional (Conv) blocks. In addition, works such as [\cite{howard2017mobilenets}], [\cite{wu2018shift}] and [\cite{chen2020addernet}] are carried out using alternatives of Conv blocks. All of these works are proven to enhance the performance of inference. 

Meanwhile, model compression techniques include quantization, pruning, knowledge distillation and low-rank decomposition. Research on quantization includes Post-Training Quantization (PTQ) [\cite{nagel2019data}], [\cite{wang2020towards}] and [\cite{cai2020zeroq}] as well as Quantization-Aware Training (QAT) [\cite{Esser2020LEARNED}] and [\cite{9837828}]. PTQ performs quantization on pre-trained models, which is free of extra calculations during training. QAT incorporates quantization error into training, making models more robust to quantization. PTQ is compatible with high-bit quantization such as 16-bit or 8-bit, whereas QAT is suitable for lower-bit quantization below 8-bit.

Structural re-parameterization, particularly used in RepVGG [\cite{ding2021repvgg}], is an effective approach for simplifying model structure while maintaining high performance for inference. As for RepVGG, it has a 3-branch structure for training and a single-branch VGG-like structure for inference. Such a decoupled structure between training and inference revives VGG again. Though the VGG-like inference structures only contain Conv and non-linear blocks, they are able to outperform ResNet and EfficientNet with fewer parameters.

\begin{figure*}[t]
    \centering
    \includegraphics[width=0.95\textwidth]{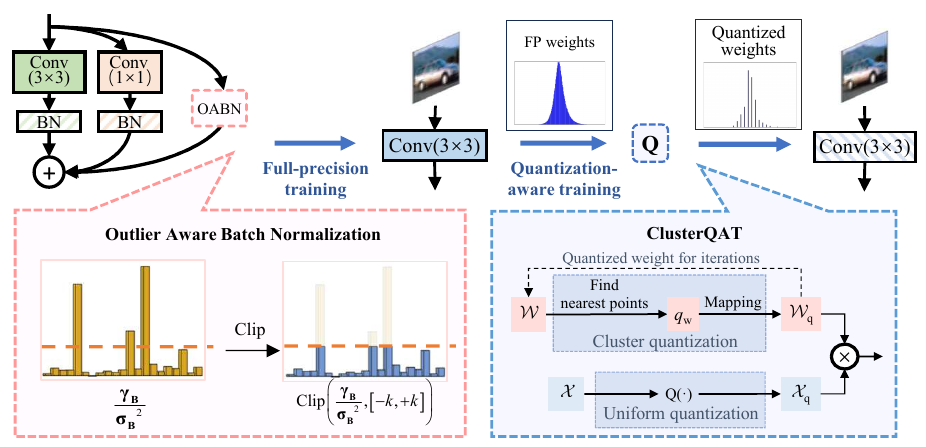}
    \caption{Overview of outlier-aware training for low-bit quantization of structural reparameterized networks.}
    \label{fig:overall}
\end{figure*}

\begin{figure}[ht]
    \centering
        \subfigure[]{
        \label{fig2a}
        \includegraphics[width=0.47\columnwidth]{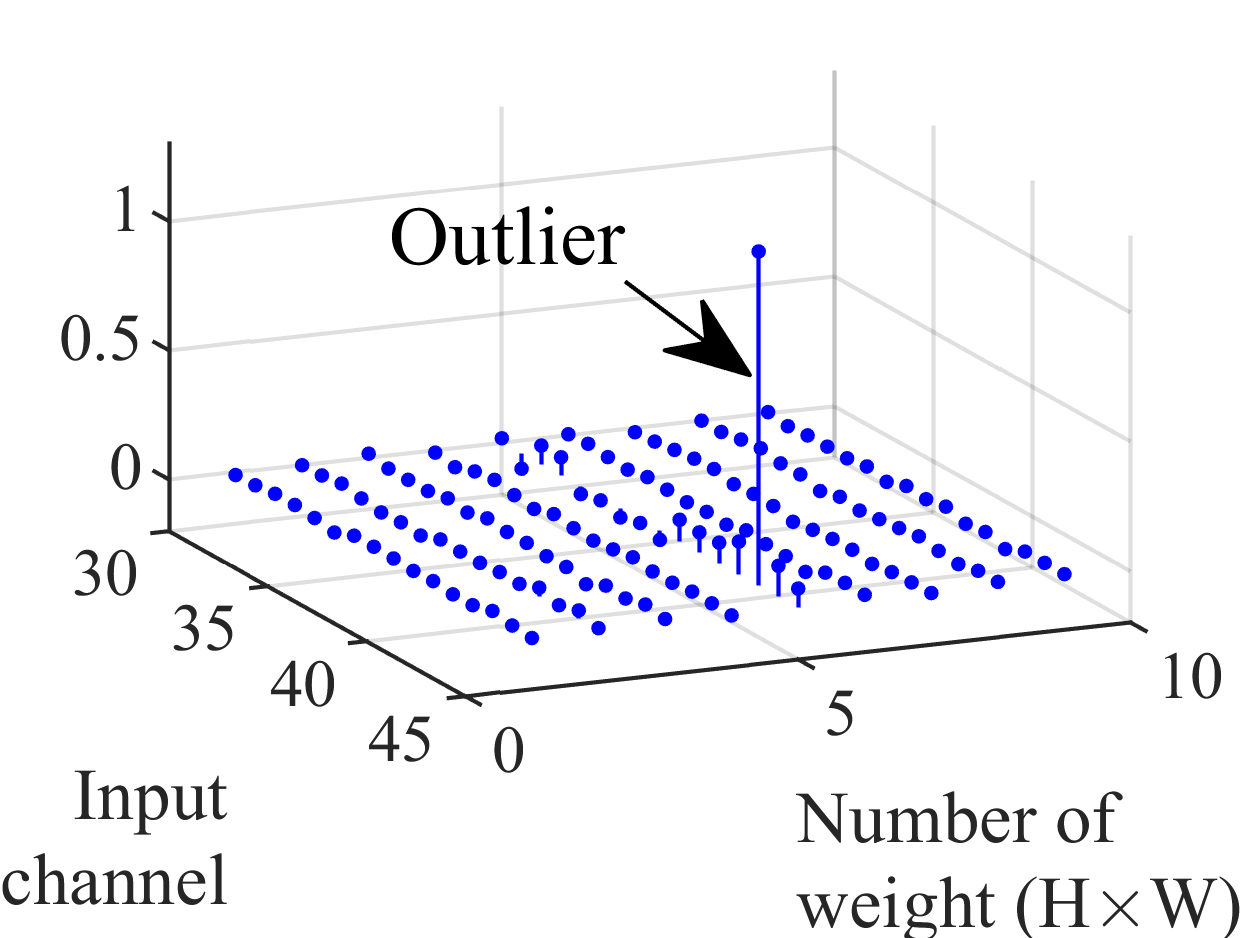}
        }
         \subfigure[]{
        \label{fig2b}
        \includegraphics[width=0.47\columnwidth]{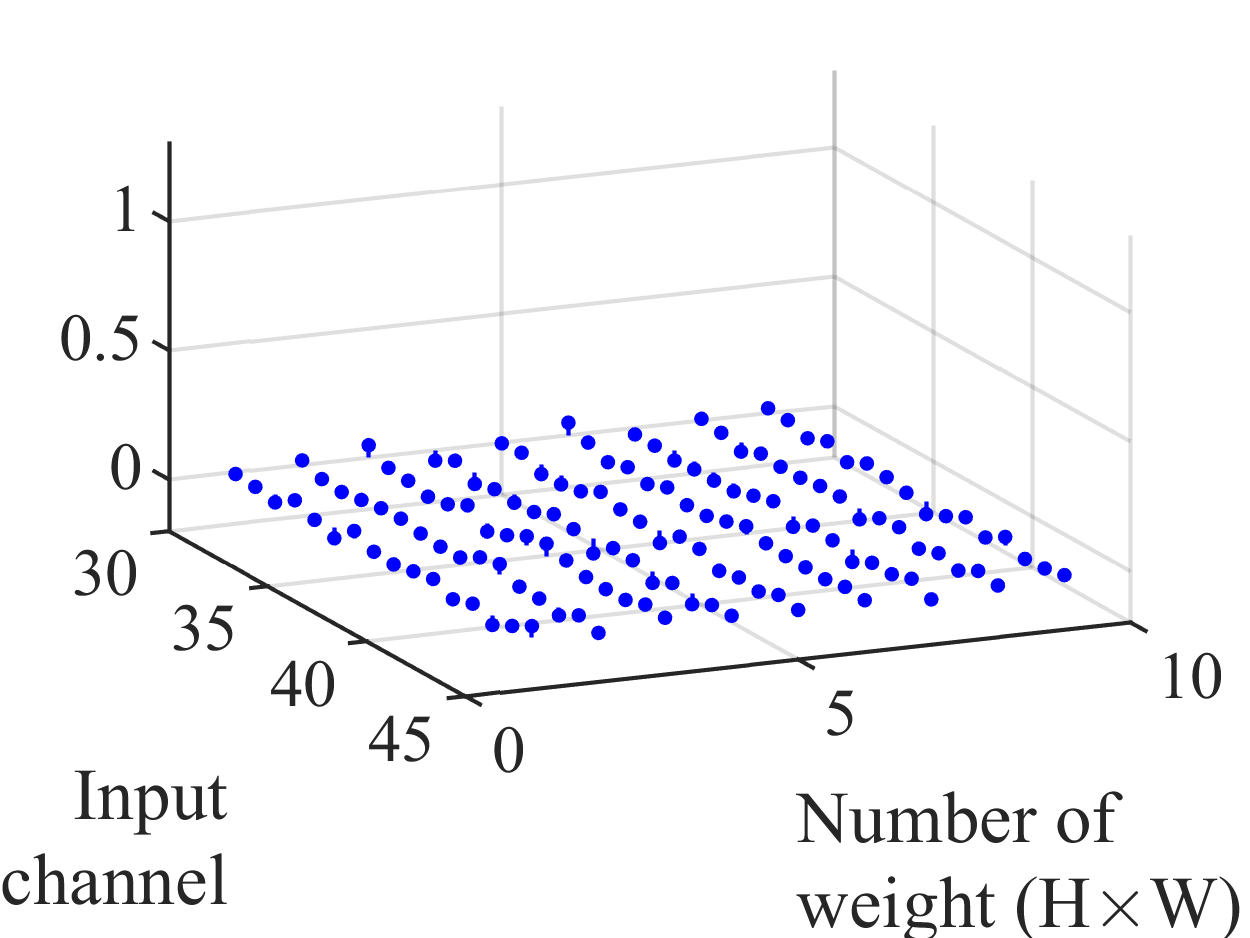}
        }
    \caption{Visualization of weights in the 43th channel of the 5th
    layer for example. (a) RepVGG - hard to quantize. (b) ResNet - easy to quantize.}
    \label{fig2 - Weights with/without Outliers}
\end{figure}

However, some of these novel architectures or operators are incompatible with techniques for model compression. As a noteworthy deficiency, RepVGG suffers from poor quantization performance. As shown in Figure \ref{fig2 - Weights with/without Outliers}, outliers exist in the weights of RepVGG. As demonstrated in [\cite{xiao2023smoothquant}], the emergence of these outliers will bring a detrimental impact on quantization.

Though previous works attempt to overcome quantization problems on different kinds of CNNs by various techniques, they fail to adapt structural re-parameterized networks. Most of them are designed and benchmarked on traditional CNNs like ResNet, MobileNet, etc. Therefore, it may be inappropriate to directly apply these techniques on RepVGG without any modifications.

There are also works on modifying RepVGG to make it more amenable to quantization, such as QARepVGG in [\cite{chu2022make}] and Rep-Optimizer in [\cite{ding2022re}]. However, these works are benchmarked only under 8-bit quantization, which is a relatively lenient condition. Discussions under stricter conditions are necessary to bring it closer to real-world hardware deployment.

Our work gives detailed discussions about the quantization of RepVGG. Apart from other works on changing model architectures or optimizers like QARepVGG and Rep-Optimizer, we propose an operator-level improvement for RepVGG by modifying its Batch-Normalization (BN) blocks during training. Then, combined with a non-uniform quantization scheme for such training techniques, we finally make RepVGG feasible for low-bit quantization. Our contributions are detailed as follows.

\begin{itemize}
    \item We identify the cause of the deteriorated quantization of RepVGG. By introducing an enhanced batch normalization operator, so-called Outlier-Aware Batch Normalization (OABN), outliers are suppressed through training. OABN enables RepVGG to be compatible with common PTQ methods, which significantly improves the quantization performance.
    \item We propose ClusterQAT, a non-uniform QAT framework for fine-tuning the models trained with OABN. By incorporating clustering into training, ClusterQAT dynamically adjusts the quantization intervals. Benefited by the preservation of distribution patterns in weights, lower-bit quantization becomes feasible on RepVGG.
    \item Combined with both PTQ and QAT, our efficient training pipeline addresses the difficulties associated with the low-bit quantization of SR networks. By carefully managing the number of states during quantization, we strike a balance between memory overhead and inference accuracy to a great extent.
\end{itemize}

\section{OABN - OPERATOR FOR IMPROVING REPVGG'S QUANTIZATION}
\subsection{REPVGG HAS OUTLIERS IN WEIGHTS}
Differing from other types of CNNs, RepVGG has synthesized weights from the merging steps between training and inference. It is the merging steps that introduce outliers into weights, which causes difficulties in quantization. 

\begin{figure}[!h]
    \centering
    \includegraphics[width=\columnwidth]{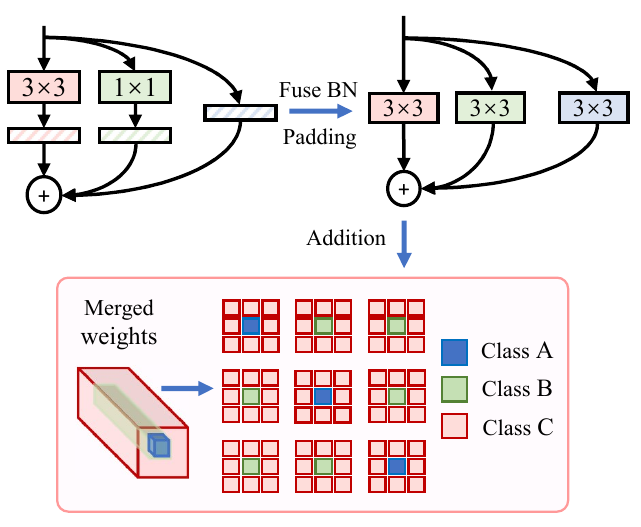}
    \caption{Merging steps in RepVGG. Each branch is converted into $3\times3$ kernel first and then added up to form the final $3\times3$ Conv kernel.}
    \label{fig10-Merging}
\end{figure}

The merging steps of RepVGG include BN fusion, padding and addition. For $3\times3$ and $1\times1$ branches, the BN blocks underpinned are firstly absorbed into Conv kernels with necessary paddings. For identity branches, ratios of $\frac{\gamma}{\sqrt{\sigma^2}}$ from BN blocks are calculated as the equivalent weights. The three branches are then added up to form the equivalent $3\times3$ Conv blocks with biases. 

Figure \ref{fig10-Merging} shows the merging procedures in detail. As for the merged weights, we define them as Class A, B and C as follows according to their positions. Those located in \textbf{Class A} are merged from all three branches. For the weights in \textbf{Class B}, they come from $3\times3$ and $1\times1$ branches, except for identity branches. The remaining weights, which are defined as \textbf{Class C}, only come from $3\times3$ branches. The quantities of the three classes in the merged weights vary from least to most.


Accordingly, the weights are computed as 
\begin{equation}
    \mathcal{W}_{\mathrm{Class\;A}} = \mathcal{W}^{(3\times3)}_{\mathrm{Class\;A}} + \mathcal{W}^{(1\times1)}_{\mathrm{Class\;A}} + 
    \mathcal{W}^{(\mathrm{id})}_{\mathrm{Class\;A}}
    \label{ClassA}
\end{equation}
\begin{equation}
    \mathcal{W}_{\mathrm{Class\;B}} = \mathcal{W}^{(3\times3)}_{\mathrm{Class\;B}} + \mathcal{W}^{(1\times1)}_{\mathrm{Class\;B}} 
\end{equation}
\begin{equation}
    \mathcal{W}_{\mathrm{Class\;C}} = \mathcal{W}^{(3\times3)}_{\mathrm{Class\;C}}
\end{equation}

\begin{figure}[h]
    \centering
        \subfigure[]{
        \includegraphics[width=0.7\columnwidth]{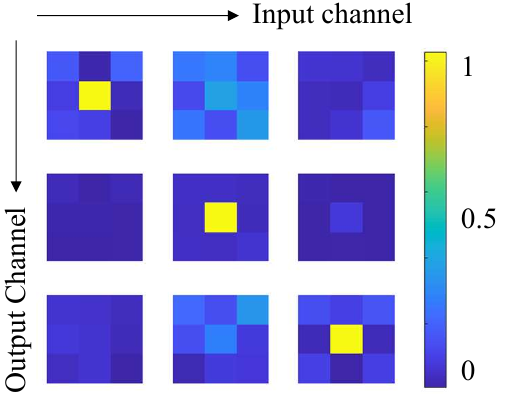}
        \label{fig12a}
        }
    \centering
        \subfigure[]{
        \includegraphics[width=\columnwidth]{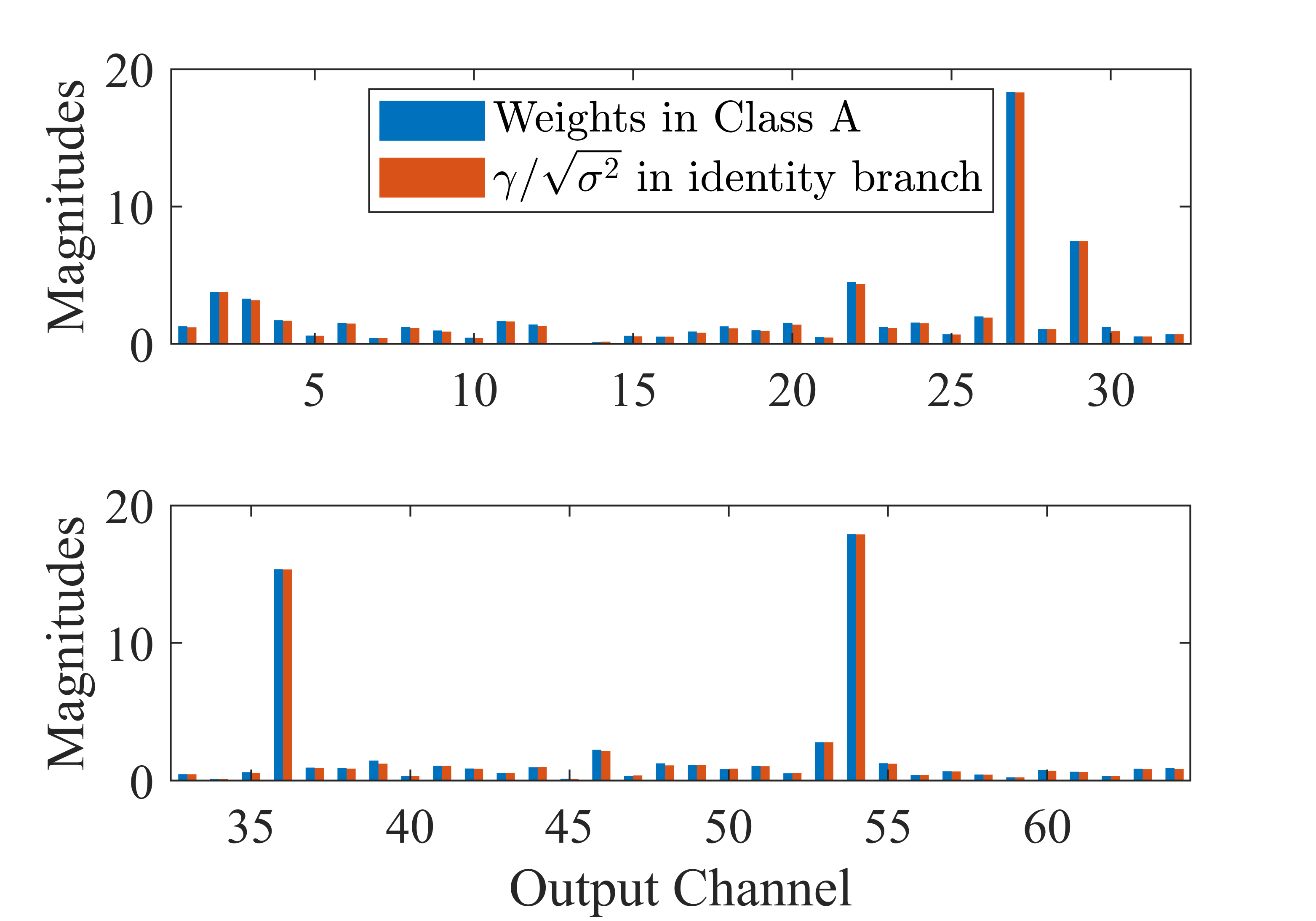}
        \label{fig12b}
        }
    \caption{Findings about positions and magnitudes of the merged weights. (a) Magnitudes of merged weights fit the pattern in Figure \ref{fig10-Merging} about the diagonal elements in Class A (e.g. in layer 2). (b) Comparisons of channel-wise ratios $\frac{\gamma}{\sqrt{\sigma^{2}}}$ $(\mathcal{W}^\mathrm{(id)}_\mathrm{Class\;A})$ and weights in Class A$(\mathcal{W}_\mathrm{Class\;A})$.}
    \label{fig12}
\end{figure}

Taking the $1^{st}$ to $3^{rd}$ output channels of the $2^{nd}$ layer into account, Figure \ref{fig12a} gives a representative magnitude pattern of the merged weights from a well-trained RepVGG. It is worth noting that extremely large weights are present on the diagonal, which are significantly different from their surroundings. By contrasting their positions and magnitudes, we reach the following findings.
\begin{itemize}
    \item \textbf{Positions} Most of the large weights are located in the position of Class A, instead of Class B and Class C.
    \item \textbf{Magnitudes} Weights located at Class A ($\mathcal{W}_{\mathrm{Class\;A}}$) are nearly equal to the $\frac{\gamma}{\sqrt{\sigma^{2}}}$ values of identity BN. $3\times3$ branches and $1\times1$ branches contribute little to Class A.
\end{itemize}
Based on the twofold findings in Figure \ref{fig12}, we confirm that it is the identity BN branches that introduce outliers. The $3\times3$ and $1\times1$ branches are not the main contributors.

\subsection{OABN - OUTLIER AWARE BATCH NORNALIZATION}
Upon identifying the outliers originating from the identity BN blocks during training, we hereby propose our OABN to remove such outliers. 

A conventional BN includes trainable parameters – $\boldsymbol{\gamma}$, $\boldsymbol{\beta}$ as well as non-trainable parameters – $\boldsymbol{\mu}$, $\boldsymbol{\sigma}$ [\cite{ioffe2015batch}]. $\boldsymbol{\mu}$ and $\boldsymbol{\sigma}$ are statistical parameters collected from mini-batches of input data. $\boldsymbol{\gamma}$ and $\boldsymbol{\beta}$ are trainable parameters calculated by accumulation of gradients. According to the merging rules in [\cite{ding2022re}], the identity branches are calculated as (\ref{Fm - Yid}).
\begin{equation}
\begin{aligned}
& \mathcal{Y}^{(\mathrm{id})}=\mathrm{BN}^{(\mathrm{id})}(\mathcal{X}) \\
& =\frac{\boldsymbol{\gamma}^{(\mathrm{id})}}{\boldsymbol{\sigma}^{(\mathrm{id})}_{\mathbf{B}}} \mathcal{X}+\left(\boldsymbol{\beta}_{\mathbf{B}}^{(\mathrm{id})}-\frac{\boldsymbol{\gamma}^{(\mathrm{id})}}{\boldsymbol{\sigma}^{(\mathrm{id})}_{\mathbf{B}}} \boldsymbol{\mu}^{(\mathrm{id})}_{\mathbf{B}}\right) \\
&
\end{aligned}
\label{Fm - Yid}
\end{equation}
We clip $\boldsymbol{\gamma}$ according to the running variance $\boldsymbol{\sigma}$ during training. The training process using OABN is given in Algorithm \ref{alg3}.

\begin{algorithm}[!h]
	\caption{Training RepVGG Block with OABN} 
	\label{alg3} 
	\begin{algorithmic}[1]
            \REQUIRE Mini-batches of training samples $\mathcal{X}$, number of mini-batches \textit{N}, scratch RepVGG block $\mathcal{M}$ with $\boldsymbol{\mu}$,$\boldsymbol{\sigma}$,$\boldsymbol{\gamma}$,$\boldsymbol{\beta}$ from identity BN, clipping threshold $\textit{k}$, number of epochs $\textit{E}_O$
		\ENSURE \textbf{VGG-like} block $\mathcal{M}_C$
            \WHILE {$\mathrm{epoch}$ \textless{} $\textit{E}_O$}
            \FOR{$n=0$ to $N-1$}
            \STATE $\boldsymbol{\mu},\boldsymbol{\sigma}^{2}$ $\leftarrow$ $\mathrm{Identity BN}(\mathcal{X})$ 
            \STATE $\boldsymbol{\gamma}_C$ $\leftarrow$ Clip($\boldsymbol{\gamma}$,[$-\textit{k}\boldsymbol{\sigma}^2$,$\textit{k}\boldsymbol{\sigma}^2$])
            \STATE Forward propagation($\mathcal{M}(\boldsymbol{\gamma}_C)$,$\mathcal{X}$)
            \STATE Backward propagation($\mathcal{M}(\boldsymbol{\gamma})$,$\mathcal{X}$)
            \STATE Update $\mathcal{M}$
            \ENDFOR
		\ENDWHILE
            \STATE $\mathcal{M}_C$ $\leftarrow$ $\mathrm{Convert\;to\;VGG}(\mathcal{M})$
	\end{algorithmic} 
\end{algorithm}


\section{CLUSTERQAT - LOWER-BIT QUANTIZATION FOR REPVGG}

In order to minimize the memory consumption of the model, joint efforts from both the model itself and algorithms for quantization are needed. Further investigation of lower-bit quantization of RepVGG based on OABN is necessary. We introduce a non-uniform quantization approach combined with clustering during training named ClusterQAT.

\subsection{QUANTIZATION-AWARE TRAINING FOR REPVGG}
QAT is the conventional way of training quantized CNN models. 
However, directly using QAT for the 3-branch RepVGG is not realizable. All trainable parameters exist in the 3-branch structure, while quantization needs to be ready for inference where single-branch weights are involved. In this case, if QAT is used directly, the quantization error cannot be directly fed back into each trainable parameter of the three-branch structure. 

To make RepVGG feasible for QAT, we quantize the merged weights during training, instead of the 3-branch weights. As shown in Figure \ref{fig:overall}, quantized weights and activations are used for forward propagation, while full precision for backward propagation.


\subsection{CLUSTERQAT - QUANTIZATION-AWARE TRAINING WITH CLUSTERING}
The key to the quality of quantization lies in whether the quantization method can preserve the original distribution of the full-precision weights. Inspired by clustering algorithms for unlabelled data, we hereby perform clustering during training for lower-bit quantization of RepVGG. As clustering could minimize the inner-class similarity and maximize the intra-class similarity as well, the proposed ClusterQAT combines such clustering quantization into QAT.

The proposed ClusterQAT perform quantization as well as updating quantization points during training. Pseudo-codes for ClusterQAT are given in Algorithm \ref{alg1}.

\begin{algorithm}[!h]
	\caption{Training RepVGG Block with ClusterQAT} 
	\label{alg1} 
	\begin{algorithmic}[1]
            \REQUIRE Mini-batches of training samples $\mathcal{X}$, pretrained 3-branch RepVGG block $\mathcal{M}$, number of mini-batches \textit{N}, number of epochs $\textit{E}_C$, bit-width for quantization \textit{b}
		\ENSURE \textbf{VGG-like} block $\mathcal{M}_C$
            \WHILE {$\mathrm{epoch} < {E}_C$}
            \STATE $q_{\mathrm{points}}^n$ $\leftarrow$ $\mathrm{linspace}(\mathrm{min}(\mathcal{M}_C),\mathrm{max}(\mathcal{M}_C),2^b)$
            \FOR{$n=0$ to $N-1$}
            \STATE $\mathcal{M}_C$ $\leftarrow$ $\mathrm{Convert\;to\;VGG}(\mathcal{M})$
            \STATE $\mathcal{M}_C$ $\leftarrow$ $\mathrm{Find\;Nearest\;Points}(\mathcal{M}_C,q_{\mathrm{points}}^n)$
            \STATE $q_{\mathrm{points}}^n$ $\leftarrow$ $\mathrm{Map}(\mathcal{M}_C,q_{\mathrm{points}}^n)$
            \STATE $\mathrm{Q}(\mathcal{X})$ $\leftarrow$ $\mathrm{Uniform\;Quantize(\mathcal{X},\textit{b})}$
            \STATE Forward Propagation ($\mathcal{M}_C$,$\mathrm{Q}(\mathcal{X})$)
            \STATE Backward Propagation ($\mathcal{M}$,$\mathcal{X}$)
            \STATE Update $\mathcal{M}$
            \ENDFOR
		\ENDWHILE
	\end{algorithmic} 
\end{algorithm}

\section{EXPERIMENTS AND DISCUSSIONS}
\subsection{EXPERIMENT SETTINGS}
We verify our proposed OABN and ClusterQAT by making comparisons with different bit-widths as well as different networks. We use NVIDIA RTX-3090 11GB for training, with software platforms of Pytorch 1.11.0 and Ubuntu 20.04. 

\begin{figure}[h]
    \centering
        \subfigure[]{
        \includegraphics[width=\columnwidth]{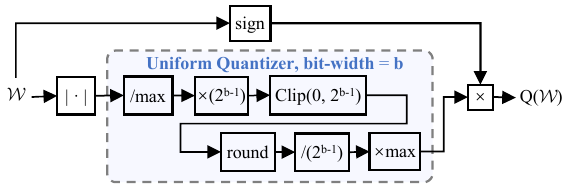}
        \label{figqweight}
        }
        \subfigure[]{
        \includegraphics[height=1.5cm,keepaspectratio]{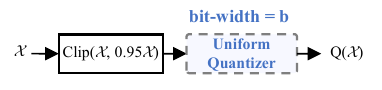}
        \label{figqact}
        }
    \caption{Uniform quantization schemes for weights and activations, respectively. (a) Weight quantizer (b) Activation quantizer.}
    \label{figquant}
\end{figure}

We first use uniform quantization to verify the effect of OABN. Specifically, weights are quantized symmetrically, while activations are quantized asymmetrically. The whole process includes scaling and rounding. A simple clamping is used to keep integer values within [0, $2^{b-1}$]. Block diagrams for uniform quantization of weights and activations are shown in Figure \ref{figquant}. 

\begin{figure}[h]
    \centering
        \subfigure[]{
        \includegraphics[width=0.45\columnwidth]{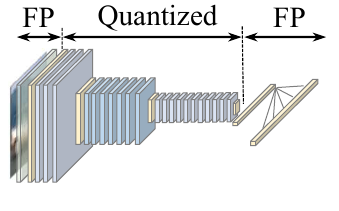}
        \label{figqweight}
        }
        \subfigure[]{
        \includegraphics[width=0.45\columnwidth]{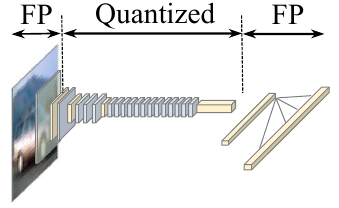}
        \label{figqact}
        }
    \caption{Quantization policies for different networks, the first layer and last layer are kept in FP32. (a) CIFAR-10 (b) ImageNet-1k.}
    \label{figpolicy}
\end{figure}

All the quantization is performed on the equivalent VGG-like Conv kernels. The first Conv layer and the last full connection (FC) layer are kept in FP32 without quantization, as shown in Figure \ref{figpolicy}. We use per-layer quantization for both weights and activations. Weights in each layer share the same quantization parameters. 

\subsection{RESULTS WITH OABN}
\subsubsection{8-bit uniform quantization with OABN}
We first verify OABN on classification datasets like CIFAR-10 [\cite{Krizhevsky09learningmultiple}] and ImageNet-1k [\cite{ILSVRC15}]. RepVGG-A0-like networks are trained with different settings of \textit{k} in Algorithm \ref{alg3}. 

The configurations are listed in Table \ref{tab:configuration}. All models are trained from scratch. The weight used for inference is selected on the epoch where FP32 accuracy reaches the top. Quantization is performed according to the rules in the appendices for both weights and activations. Figure \ref{fig_CIFAR10}, and \ref{fig_ImageNet} give the results of accuracy under FP32 and INT8.

\begin{table}[!h]
\centering
\caption{Configurations of network structure.}
\label{tab:configuration}
\begin{tblr}{
  cells = {c},
  hline{1} = {-}{0.08em},
  hline{2,4} = {-}{},
}
Dataset     & \# Blocks & \# Channels    & {\# Params.\\(M)} \\
CIFAR-10    & 4,8,12,1  & 64,64,64,64    & 1.036             \\
ImageNet-1k & 2,4,14,1  & 48.96,192,1280 & 8.309             
\end{tblr}
\end{table}

\begin{table}[h]
\centering
\caption{Configurations of hyper-parameters.}
\label{tab:Hyper}
\begin{tblr}{
  column{1} = {c},
  column{2} = {c},
  column{3} = {c},
  hline{1} = {-}{0.08em},
  hline{2,8} = {-}{},
}
{Item\textbackslash{}Dataset} & CIFAR-10            & ImageNet-1k                                          \\
Batch size                     & 128                                 & 256                                                      \\
Epochs ($\textit{E}_O$)        & 450                              & {120 ($\textit{k}\geq$ 5)\\160 (\textit{k} \textless{} 5)}         \\
Optimizer                      & SGD                                & SGD                                                      \\
Learning rate                  & 0.1                                 & 0.1                                                      \\
Scheduler                      & {CosineAnnealing}              & StepLR                                                   \\
Argument                      & $T_{max}$ = 50                & {\textit{S}\footnotemark[1] = 30 ($\textit{k}\geq$ 5)\\\textit{S} = 40 (\textit{k \textless{}} 5)} 
\end{tblr}
\end{table}
\vspace{16pt}
\footnotetext[1]{Step size is defined in short as \textit{S}.}
\begin{figure}[!h]
    \centering
        \subfigure[]{
        \includegraphics[width=1.02\columnwidth]{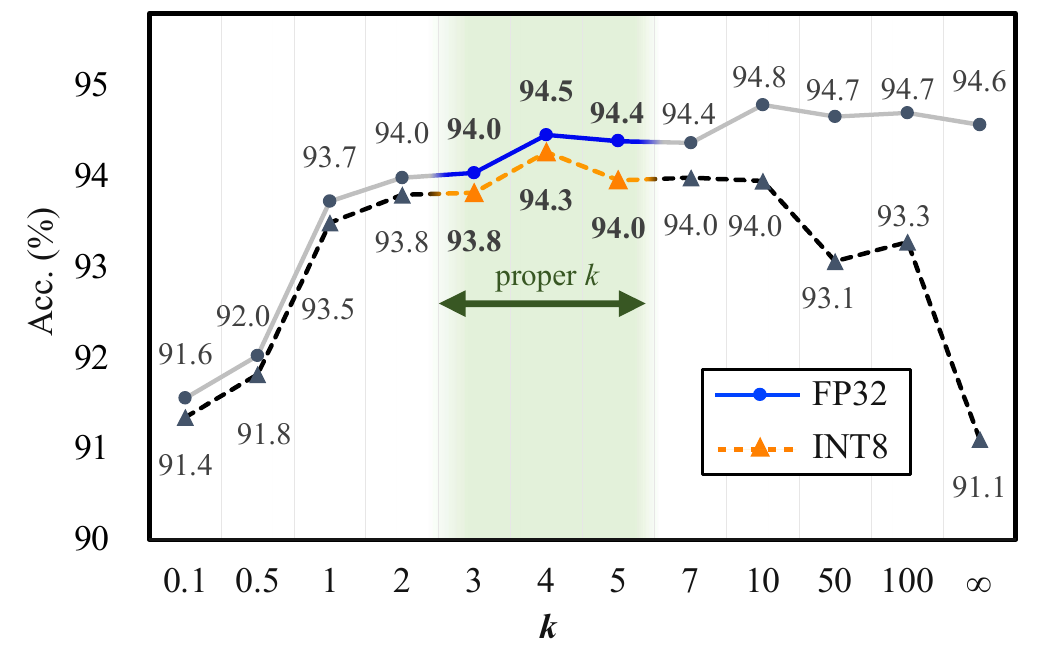}
        \label{fig_CIFAR10}
        }
        \subfigure[]{
        \includegraphics[width=1.04\columnwidth]{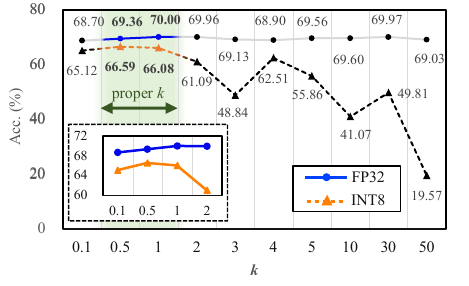}
        \label{fig_ImageNet}
        }
    \caption{Results for INT8 classification with OABN. (a) CIFAR-10 (b) ImageNet-1k.}
    \label{fig:enter-label}
\end{figure}
All plots exhibit consistent trends in accuracy between FP32 and INT8. Smaller \textit{k} bridges the gap between FP32 and INT8, whereas larger \textit{k} results in an expanded gap. For the results on CIFAR-10 presented in Figure \ref{fig_CIFAR10}, the gap in inference accuracy between FP32 and INT8 is bridged when \textit{k} $<$ 6. A smaller \textit{k} limits the ratio of $\frac{\gamma}{\sqrt{\sigma^{2}}}$ in identity BN. According to the merging rules outlined in Figure \ref{fig:overall}, the merged weights at Class A are suppressed. Correspondingly, those points expected to have relatively large magnitudes are no longer considered outliers. Quantization error is not a factor that affects the accuracy anymore. 

However, when \textit{k} is tiny (e.g. \textit{k} $<$ 1), the accuracy decreases for both FP32 and INT8. Although the gap between them is small, the accuracy of FP32 and INT8 are generally lower. Since the model's performance has already degraded, it is natural for the quantization performance also to be poor. A proper range for \textit{k} exists, which should neither be too big nor too small to ensure a successful quantization on RepVGG.

As for classification results on ImageNet shown in Fig 11(b), the curves of INT8 show similar trends compared with CIFAR-10. A relatively better \textit{k} is around 1 to 3. All of these results demonstrate the effect of OABN for training a quantization-friendly RepVGG. 

\subsubsection{8-bit PTQ results with OABN}

In this section, we pick up some PTQ techniques for quantizing our RepVGG with OABN. To examine the effectiveness, as well as the ability of these PTQ techniques, we list out the accuracy (Acc.) and the total number of unique elements for a clear demonstration. The first and last layers are free of quantization according to Figure \ref{figpolicy}. We add up the summed amount of unique weights from the remaining layers, which is defined as \textbf{state size}. Table \ref{tab:my-table} shows detailed information. All of the items are tested on ImageNet-1k.
\begin{table*}[ht]
\centering
\caption{8-bit PTQ results on ImageNet-1k.}
\label{tab:my-table}
\begin{tabular}{ccccccccccc} 
\hline
\multirow{2}{*}{\begin{tabular}[c]{@{}c@{}}Quantization settings \\ \textbackslash{} Scratch network\end{tabular}} & \multicolumn{2}{c}{\multirow{2}{*}{RepVGG-A0}}                                                                & \multicolumn{6}{c}{RepVGG-A0 (OABN)}                                                                                                                                                                                                                                                                                                          & \multicolumn{2}{c}{\multirow{2}{*}{QARepVGG-A0}}                                                               \\ 
\cline{4-9}
                                                                                                                   & \multicolumn{2}{c}{}                                                                                          & \multicolumn{2}{c}{\textit{k} = 50}                                                                           & \multicolumn{2}{c}{\textit{k} = 10}                                                                           & \multicolumn{2}{c}{\textit{k} = 0.5}                                                                          & \multicolumn{2}{c}{}                                                                                           \\ 
\hline
Method for PTQ                                                                                                     & \begin{tabular}[c]{@{}c@{}}State\\ size(k)\end{tabular} & \begin{tabular}[c]{@{}c@{}}Acc.\\ (\%)\end{tabular} & \begin{tabular}[c]{@{}c@{}}State\\ size(k)\end{tabular} & \begin{tabular}[c]{@{}c@{}}Acc.\\ (\%)\end{tabular} & \begin{tabular}[c]{@{}c@{}}State\\ size(k)\end{tabular} & \begin{tabular}[c]{@{}c@{}}Acc.\\ (\%)\end{tabular} & \begin{tabular}[c]{@{}c@{}}State\\ size(k)\end{tabular} & \begin{tabular}[c]{@{}c@{}}Acc.\\ (\%)\end{tabular} & \begin{tabular}[c]{@{}c@{}}State\\ size(k)\end{tabular} & \begin{tabular}[c]{@{}c@{}}Acc.\\ (\%)\end{tabular}  \\ 
\hline
FP32                                                                                                                 & 6987.44                                                 & 69.57                                               & 6987.48                                                 & 69.03                                               & 6987.74                                                 & 69.60                                               & 6986.94                                                 & 69.36                                               & 6984.50                                                 & 70.04                                                \\ 
\hline
Uniform (Ours)                                                                                                     & 2.56                                                    & 43.09                                               & 2.56                                                    & 19.57                                               & 2.69                                                    & 41.07                                               & \textbf{1.83}                                           & \textbf{66.59}                                      & 2.90                                                    & 64.78                                                \\
DFQ                                                                                                                & 3.70                                                    & 42.78                                               & 3.74                                                    & 18.66                                               & 3.89                                                    & 40.54                                               & 4.21                                                    & 66.53                                               & 3.94                                                    & 64.72                                                \\
AdaQuant                                                                                                           & 399.19                                                  & 67.70                                               & 400.93                                                  & 67.20                                               & 403.20                                                  & 68.08                                               & 507.14                                                  & 68.20                                               & 373.99                                                  & 69.65                                                \\
ZeroQ                                                                                                              & 398.69                                                  & 67.33                                               & 400.51                                                  & 66.57                                               & 402.84                                                  & 66.96                                               & 506.55                                                  & 68.55                                               & 373.79                                                  & 69.12                                                \\
BitSplit                                                                                                           & \textasciitilde{}                                       & 0.00                                                & \textasciitilde{}                                       & 0.00                                                & \textasciitilde{}                                       & 0.10                                                & 452.10                                                  & 69.32                                               & 294.77                                                  & 69.87                                                \\
\hline
\end{tabular}
\end{table*}

It is obvious that OABN significantly enhances RepVGG’s PTQ performance under 8-bit settings. All approaches yield improved results for RepVGG when OABN is applied. Specifically, for small values of \textit{k} (e.g. \textit{k} = 0.5), all PTQ methods mentioned above become effective, while they are less effective for larger values of \textit{k}.

However, some of the quantization methods are not able to compress the state size. They produce 100$\times$ larger state size compared with our uniform quantization, which needs extra memory for inference. From this perspective, our uniform quantization is advantageous as it holds relatively fewer states while keeping acceptable accuracy under 8-bit settings.

Compared with QARepVGG, we can find that QARepVGG performs well under loose quantization conditions where more states of weights are produced. Regarding rigorous conditions with fewer states of weights, our RepVGG with OABN outperforms QARepVGG.
\subsection{RESULTS WITH OABN AND CLUSTERQAT}
\subsubsection{Lower-bit Quantization with OABN and ClusterQAT}

Based on the pre-trained RepVGG models using OABN, we finetune our models using QAT according to the procedures given in Table \ref{tab:ConfigOACL}. 

\begin{table}[!h]
\centering
\caption{Training pipelines with OABN and ClusterQAT.}
\label{tab:ConfigOACL}
\begin{tabular}{cccc} 
\hline
\multirow{2}{*}{Epochs}                 & \multicolumn{3}{c}{Quantization Settings}  \\ 
\cline{2-4}
                                        & Cluster & Quantize & OABN              \\ 
\cline{1-4}
from 0 to 20                            & False   & False        & False             \\
from 20 to $\textit{E}_O$                & False   & False        & True              \\
from $\textit{E}_O$ to $\textit{E}_O+\textit{E}_C$ & True    & True         & False             \\
\hline
\end{tabular}
\end{table}

\begin{figure}[h]
    \centering
    \includegraphics[width=\columnwidth]{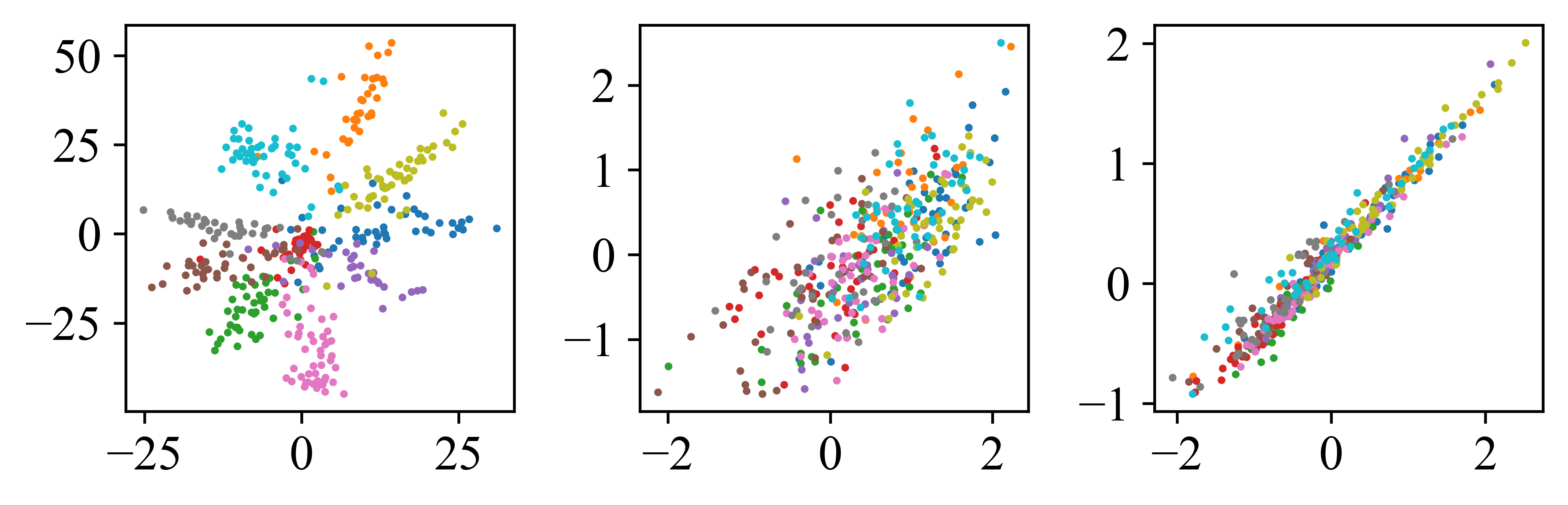}
    \caption{Efforts of OABN and ClusterQAT under 4-bit quantization. ClusterQAT must be combined with OABN. (Left) CluserQAT with OABN (\textit{k}=0.5). (Middle) ClusterQAT with traditional BN. (Right) uniform QAT with OABN. }
    \label{fig-jointOABNClusterQAT}
\end{figure}

Figure \ref{fig-jointOABNClusterQAT} shows an ablation study about the joint efforts with OABN and ClusterQAT under 4-bit quantization. It could be seen that ClusterQAT makes an effort only for models trained with OABN. Directly using ClusterQAT on models that are trained without OABN will not help under low-bit quantization.

Furthermore, more trials are provided to verify our ClusterQAT. To make a clear demonstration, we fine-tune some representative models trained with different \textit{k} using OABN. Comparisons with uniform QAT, as well as the cases under different s are listed. Results on CIFAR-10 and ImageNet-1k are given in Table \ref{tab:ClusterQATCF10} and Table \ref{tab-ClImageNet}, respectively. All the quantization are performed \textbf{layer-wise}, which aligns with the policies used Figure \ref{fig:enter-label}.

\footnotetext[4]{We define \textbf{the method of quantization} as quant. in short.}

\begin{table*}[ht]
\centering
\caption{Results of ClusterQAT on CIFAR-10.}
\label{tab:ClusterQATCF10}
\begin{tabular}{ccccccccc} 
\hline
\multicolumn{4}{c}{Quantization Settings\textbackslash{} Pre-trained Network}                                                 & \multirow{3}{*}{RepVGG-A0} & \multicolumn{3}{c}{RepVGG-A0 (OABN)}                 & \multirow{3}{*}{QARepVGG-A0}  \\ 
\cline{1-4}\cline{6-8}
Act.  & \begin{tabular}[c]{@{}c@{}}Act. \\ quant.\protect\footnotemark[4]\end{tabular} & Wgt. & \begin{tabular}[c]{@{}c@{}}Wgt.\\ quant.\end{tabular} &                            & \textit{k} = 100 & \textit{k} = 3 & \textit{k} = 0.5 &                               \\ 
\hline
\multicolumn{4}{c}{FP32}                                                                                                        & 94.57                      & \textbf{94.7}    & 94.04          & 92.03            & 94.3                          \\ 
\hline
8     & \multirow{4}{*}{Uniform}                               & 8    & \multirow{4}{*}{Uniform}                              & 92.28                      & \textbf{94.25}   & 93.88          & 91.98            & 94.06                         \\
4     &                                                        & 5    &                                                       & 10.00                      & 11.90            & 80.09          & \textbf{88.74}   & 10.68                         \\
4     &                                                        & 4    &                                                       & 10.00                      & 10.00            & 10.00          & 10.10            & 11.60                         \\
4     &                                                        & 3    &                                                       & 10.00                      & 10.00            & 10.00          & 10.66            & 10.26                         \\ 
\hline
8     & \multirow{4}{*}{Uniform}                               & 8    & \multirow{4}{*}{Cluster}                              & 93.92                      & \textbf{94.56}   & 93.74          & 91.92            & 94.19                         \\
4     &                                                        & 5    &                                                       & 11.78                      & 85.88            & \textbf{89.89} & 89.05            & 88.80                         \\
4     &                                                        & 4    &                                                       & 11.76                      & 39.36            & \textbf{89.46} & 88.52            & 88.24                         \\
4     &                                                        & 3    &                                                       & 10.01                      & 13.64            & 10.58          & \textbf{88.05}   & 13.87                         \\
\hline
\end{tabular}
\end{table*}

\begin{table*}[ht]
\centering
\caption{Results of ClusterQAT on ImageNet-1k.}
\label{tab-ClImageNet}
\begin{tabular}{clccccccc} 
\hline
\multicolumn{4}{c}{Quantization settings\textbackslash{} Pre-trained network}                                                                      & \multirow{3}{*}{RepVGG-A0} & \multicolumn{3}{c}{RepVGG-A0 (OABN)}                & \multirow{3}{*}{QARepVGG-A0}  \\ 
\cline{1-4}\cline{6-8}
Act.  & \multicolumn{1}{c}{\begin{tabular}[c]{@{}c@{}}Act. \\ quant.\end{tabular}} & Wgt.  & \begin{tabular}[c]{@{}c@{}}Wgt.\\ quant.\end{tabular} &                            & \textit{k} = 50 & \textit{k} = 4 & \textit{k} = 0.5 &                               \\ 
\hline
\multicolumn{4}{c}{FP32}                                                                                                                             & 69.57                      & 69.03           & 68.90          & 69.36            & \textbf{70.04}                \\ 
\hline
8     & \multicolumn{1}{c}{Uniform}                                                & 8     & Uniform                                               & 43.09                      & 19.57           & 62.51          & \textbf{66.59}   & 64.78                         \\ 
\hline
8     & \multirow{4}{*}{Uniform}                                                   & 8     & \multirow{4}{*}{Cluster}                              & 67.85                      & 18.12           & 64.32          & 69.10            & \textbf{69.96}                \\
7     &                                                                            & 7     &                                                       & 0.29                       & 20.48           & 57.05          & \textbf{68.73}   & 60.01                         \\
6     &                                                                            & 6     &                                                       & 28.05                      & 27.21           & 34.61          & 67.29            & \textbf{68.18}                \\
5     &                                                                            & 5     &                                                       & 28.20                      & 0.12            & 34.23          & \textbf{63.68}   & 60.59                         \\
\hline
\end{tabular}
\end{table*}

As shown in Table \ref{tab:ClusterQATCF10}, our proposed ClusterQAT outperforms traditional uniform QAT in terms of performance on the CIFAR-10 dataset. However, it requires cooperation with a pre-trained model using an appropriate \textit{k} from OABN. Specifically, for the RepVGG-A0 model trained without OABN, results are unacceptable even when using ClusterQAT. The error introduced by outliers is so large that ClusterQAT fails to handle. Remarkably, when the clipping threshold $\textit{k}$ is adjusted to its optimal value (e.g. 0.5), our model achieves good accuracy again even at a of 3. ClusterQAT shows improvement compared to models trained with larger \textit{k}. In contrast, for all trials using uniform quantization, adjusting $\textit{k}$ does not significantly impact performance. We observe similar trends on the ImageNet dataset, as shown in Table \ref{tab-ClImageNet}.

\vspace{2pt}
 \subsection{COMPUTATION COST FOR TRAINING}
The training speed is benchmarked using single RTX-3090 11GB. As shown in Table \ref{tab:TrainingCost}, our RepVGG+OABN consumes less pass size and FLOPs compared with QARepVGG.

\begin{table}[h]
\centering
\caption{Time and computation costs for FP32 training on ImageNet-1k.}
\label{tab:TrainingCost}
\begin{tabular}{cccc} 
\hline
Network                                                    & \begin{tabular}[c]{@{}c@{}}\# Trainable \\ params.(M)\end{tabular} & \begin{tabular}[c]{@{}c@{}}Pass\\~size(MB)\end{tabular} & GFLOPs          \\ 
\hline
RepVGG-A0                                                  & 9.108                                                              & 88.740                                                  & 1.530           \\
\begin{tabular}[c]{@{}c@{}}RepVGG-A0\\~(OABN)\end{tabular} & \textbf{9.103}                                                     & \textbf{88.740}                                         & \textbf{1.527}  \\
QARepVGG-A0                                                & 9.108                                                              & 95.820                                                  & 3.054           \\
\hline
\end{tabular}
\end{table}
\section{CONCLUSIONS}
RepVGG is a novel CNN architecture that revives VGG again. However, its merging step introduces outliers and uniquely distributed weights for inference, making it unfriendly to quantization. By using OABN instead of traditional BN during training, the well-trained RepVGG are free of outliers, which is friendly to 8-bit quantization. Based on the models trained by OABN, we propose ClusterQAT to adjust the weights further. Integrating OABN and ClusterQAT, we make RepVGG friendly to state-size limited quantization.

\section{RELATED WORK}
QARepVGG improves the quantization of RepVGG from another perspective. It believes that it is the large variance of weights and activations that hinder the performance of INT8 quantization. To solve the problem, a new training structure with a post-BN block is proposed. Results show that classification accuracy on ImageNet under INT8 quantization rises 20\% compared with RepVGG.


\printbibliography

\end{document}